\documentclass[preprint]{elsarticle}

\pdfoutput=1	

\usepackage{rotating} 
\usepackage{booktabs} 
\usepackage{wrapfig} 
\usepackage{subcaption} 
\usepackage{makecell} 
\usepackage{amsmath}
\usepackage{amsfonts}
\usepackage{amssymb} 
\usepackage{bm} 
\usepackage{multicol}

\usepackage{amsthm} 

\theoremstyle{definition} 
\newtheorem{proposition}{Proposition}

\theoremstyle{remark}

\newcommand{\me}{\mathrm{e}} 

\begin{document}

\begin{frontmatter}

\title{A Deep Learning Interpretable Classifier for Diabetic Retinopathy Disease Grading}

\author[rvt]{Jordi de la Torre\corref{cor1}}
\ead{jordi.delatorre@gmail.com}
\author[rvt]{Aida Valls}
\ead{aida.valls@urv.cat}
\author[rvt]{Domenec Puig}
\ead{domenec.puig@urv.cat}

\cortext[cor1]{Corresponding author}

\address[rvt]{Departament d'Enginyeria Inform\`atica i Matem\`atiques.\\Escola T\`ecnica Superior d'Enginyeria.\\Universitat Rovira i Virgili\\Avinguda Paisos Catalans, 26. E-43007\\
Tarragona, Spain}

\date{Dec 21, 2017}

\begin{abstract}
Deep neural network models have been proven to be very successful in image classification tasks, also for medical diagnosis, but their main concern is its lack of interpretability. They use to work as intuition machines with high statistical confidence but unable to give interpretable explanations about the reported results. The vast amount of parameters of these models make difficult to infer a rationale interpretation from them. In this paper we present a diabetic retinopathy interpretable classifier able to classify retine images into the different levels of disease severity and of explaining its results by assigning a score for every point in the hidden and input space, evaluating its contribution to the final classification in a linear way. The generated visual maps can be interpreted by an expert in order to compare its own knowledge with the interpretation given by the model. 
\end{abstract}


\end{frontmatter}


\section{Introduction}

Deep Learning methods have been used extensively in the last years for many automatic classification tasks. For the case of image analysis, the usual procedure consists on extracting the important features with a set of convolutional layers and, after that, make a final classification with these features using a set of fully connected layers. Finally, a soft-max output layer gives as a result the predicted output probabilities of the set of classes predefined in the model. During training, model parameters are changed using a gradient-based optimization algorithm, which minimizes a predefined loss function. 

Once the classifier has been trained (i.e. the parameters of the different layers of the model have been fixed), the quality of the classification outputs predicted is compared against the correct "true" values stored on a labeled dataset. This data is considered as the gold standard, ideally coming from the consensus of the knowledge of a human experts committee.

This mapping allows the classification of multidimensional objects into a small number of categories. The model is composed by many neurons that are organized in layers and blocks of layers, piled together in a hierarchical way. Every neuron receives the input from a predefined set of neurons. Every connection has a parameter that corresponds to the weight of the connection. 

The function of every neuron is to make a transformation of the received inputs into a calculated output value. For every incoming connection, the weight is multiplied by the input value received by the neuron and the aggregated value is feeded to an activation function that calculates the output of the neuron. The parameters are usually optimized using a stochastic gradient descent algorithm that minimizes a predefined loss function. The parameters of the network are updated after backpropagating the loss function gradients through the network. These hierarchical models are able to learn multiple levels of representation that correspond to different levels of abstraction, which enables the representation of complex concepts in a compressed way \cite{nature-deep-learning}, \cite{888}, \cite{Bengio:2013:RLR:2498740.2498889}, \cite{bengio-2009}.
 
Deep Learning based models have been proven to be very effective when trained with enough labelled data (order of magnitude of tens of thousands of examples per class) but their main concern is its \emph{lack of interpretability}. Every successful model tend to have millions of parameters, making difficult to get from them a rationale interpretation. 

In medical diagnosis tasks is important not only the accuracy of the predictions but also the reasons behind a decision. Self-explainable models enable the physicians to contrast the information reported by the model with their own knowledge, increasing the information and the probability of a good diagnostic.  

Different attempts have been done in order to interpret the results reported by neural networks. In \cite{zeiler2014visualizing} a network propagation technique is used for the visualization of the features in the input space. After this \cite{bach2015pixel} used a pixel-wise decomposition for classification decision. This decomposition could be done in two ways: considering the network as a global function, disregarding its topology (functional approach) or using the natural properties of decomposition of the inherent topology of the function to use a message passing technique for propagating back into the pixel space the decomposition. After this, in \cite{montavon2017explaining} they used a so named Deep Taylor decomposition technique to replace the inherently intractable standard Taylor decomposition using a multitude of simpler analytically tractable Taylor decompositions.

In our work we use a similar approach to the used in the pixel-wise decomposition, taking into account the compositional nature of the topology as in \cite{zeiler2014visualizing} and \cite{bach2015pixel}. The novel approach comes from the fact that being the score globally conservative, the conservation does not hold between layers. The concept of \emph{score} in our paper is similar to the concept of \emph{relevance} used in layer-wise relevance propagation. Apart from the input-space contribution, there is also another one coming from every layer that is independent from the input-space and that depends only on the parameters of each layer. It is not an attribute of the individual pixels that has to be back-propagated but a contribution of the receptive field (RF) that represents the layer as an individual entity. We only propagate back the part of the score that depends on the precedent input for every layer. In our model explanation we consider the constant part as a property of the RF of every layer. This approach, allows us to do an exact propagation of the scores using a deconvolutional approach. Differing also from \cite{zeiler2014visualizing}, our method allows the integration of the batch normalization and of other typical neural network block constituents into the score propagation. A full set of score propagation blocks with the more typical deep learning functional constituents is derived in order to make as easy as possible the porting of the paper results to other networks and applications.

This interpretation model is tested in our application research area: diabetic retinopathy (DR). DR is a leading disabling chronic disease  and  one of the main causes of blindness and visual impairment in developed countries for diabetic patients. Studies reported that 90\% of the cases can be prevented through early detection and treatment. Eye screening through retinal images is used by physicians to detect the lesions related with this disease. Due to the increasing number of diabetic people, the amount of images to be manually analyzed is becoming unaffordable. Moreover, training new personnel for this type of image-based diagnosis is long, because it requires to acquire expertise by daily practice. Medical community establishes a standardized classification based on four severity stages \cite{diaclass} determined by the type and number of lesions (as micro-aneurysms, hemorrhages and exudates) present in the retine: class 0 referring to no apparent retinopathy, class 1 as a Mild Non-Proliferative Diabetic Retinopathy (NPDR), class 2 as Moderate NPDR, class 3 as a Severe NPDR and class 4 as a Proliferative DR. 

We design a \emph{DR interpretable image classification model} for grading the level of disease. This model is able to not only report the predicted class but also to score the importance of every pixel of the input image in the final classification decision. In such a way is possible to determine which pixels in the input image are more important in the final decision and facilitate the human experts an explanation to verify the results reported by the model.

The paper is structured as follows: in Section \ref{sec:related} the current work on deep learning applied to DR is briefly introduced, then, the main works on interpretation of DL are presented. Section \ref{sec:math} we present the complete mathematical formulation of our interpretable model describing the score propagation model, Section \ref{sec:class} describes the DR DL classification model, Section \ref{sec:results} present the results showing a set of samples of the type of visual interpretations and finally Section \ref{sec:conclusions} present the final conclusions of our work.

\section{Related Work}\label{sec:related}

Many deep learning based DR classifiers has been published in the last years. In \citep{DELATORRE2017} a deep learning classifier was published for the prediction of the different disease grades. This model was trained using the public available EyePACS dataset. The training set had 35,126 images and the test set 53,576. The quadratic weighted kappa (qwk) evaluation metric \citep{cohen1968weighted} over the test set using a unique deep learning model without ensembling was close to the reported by human experts. 

In \citep{doi:10.1001/jama.2016.17216} a deep learning classifier was published for the detection of the most severe cases of DR (grouping for the joined detection of the classes of referable DR, defined as moderate or worse DR or referable macular edema). This model was trained using an extended version of the EyePACS dataset mentioned before with a total of 128,175 images and improving the proper tagging of the images using a set of 3 to 7 experts chosen from a panel of 54 US expert Ophtalmologists. This model surpassed the human expert capabilities, reaching at the final operating point approximately  97\% sensitivity and 93.5\% specificity in the test sets of about 10,000 images for detecting the worse cases of DR. The strength of this model was its ability to predict the more severe cases with a sensitivity and specificity greater than human experts. The drawback, as many deep learning based models, is its lack of interpretability. The model acts like a \emph{intuition machine} with a highly statistical confidence but lacking an interpretation of the foundations of the final decisions making difficult to the experts to balance and compare its prior knowledge with the reasons behind the final conclusion to get even better diagnostics.

In last years different approximations have been derived to convert the initial deep learning black box classifiers into \emph{interpretable classifiers}. In the next sections we introduce the more successful interpretation models existing today: sensitivity maps, layer-wise relevance propagation and Taylor type decomposition models. 

\subsection{Sensitivity maps}

Sensitivity maps \citep{DBLP:journals/corr/SimonyanVZ13} are pixel-space matrices obtained from the calculation of $ \frac{\partial f(I)}{\partial I_{c,i,j}} \quad \forall c,i,j$. This matrices are easy to calculate for deep neural networks. They use the same backpropagation rules that are used during training, requiring only one more backpropagation step for reaching the input space. The problem with this approach is that there is no direct relationship between $f(I)$ and $\nabla f(I)$. The main concern of this models is that being the objective to explain $f(x)$, $ \frac{\partial f(I)}{\partial I_{c,i,j}}$ is only giving us information about the local change of the function. For high non-linear functions like deep neural networks the local variation is pointing to the nearest local optimum that not necessarily should be in the same direction that the global minimum \cite{baehrens2010explain}.

\subsection{Layer-wise relevance propagation} 

In \cite{bach2015pixel} the authors split the total score of a classification into individual \emph{relevance scores} that act as a positive or negative contributions to the final result. 

The method has the next general constraints: the first one is the nature of the classification function that has to be decomposable into several layers of computation (like a deep neural network), the second one that the total relevance must be preserved from one layer to another, that is to say that the relevance of one layer equals the ones of all other layers (eq. \ref{eq:relevance-conservation}) and finally that the relevance of every node must be equal to the sum of all the relevance messages incoming to such a node and also equal to the sum of all relevance messages outgoing from the same node (eq. \ref{eq:relevance-message-value}).

\begin{equation}
	f(x) = \sum_{d \in l+1}R_d^{(l+1)} = \sum_{d \in l}R_d^{(l)} = ... = \sum_{d}R_d^{(1)}
	\label{eq:relevance-conservation}
\end{equation}

\begin{equation}
R_{i \leftarrow k}^{(l,l+1)} = R_k^{(l+1)} \frac{a_i \omega_{ik}}{\sum_{h} a_h \omega_{hk}}
\label{eq:relevance-message-value}
\end{equation}

As the authors explain in \cite{bach2015pixel}, these constraints does not assure a unique way of splitting the score into the different nodes and does not guarantee the final score distribution to have a meaningful interpretation of the classifier prediction.

\subsection{Taylor-type decomposition} 

Another way for solving the interpretability problem is using the gradient of the classification function for the calculation of the next Taylor approximation \cite{bach2015pixel}:

\begin{equation}
f(I) \approx f(I_0) + \nabla(I_0) [ I - I_0] = f(I_0) + \sum_{c=1}^C \sum_{i=1}^{H} \sum_{j=1}^W \frac{\partial f}{\partial I_{c,i,j}}(I_{c,i,j} - I_{0 c, i, j}) 
\label{eq:taylor}
\end{equation}

Being $I_0$ a free parameter that should be chosen in a way that $f(I_0) = 0$ in the case of f(I) defined as a function that reports a value greater than one when belongs to the class and lower than 0 otherwise. Defined in such a way, $f(I) = 0$ express the case of maximum uncertainty about the image. Finding $I_0$ allows us to express $f(I)$ as:

\begin{equation}
f(I) \approx \nabla(I_0) [ I - I_0] = \sum_{c=1}^C \sum_{i=1}^{H} \sum_{j=1}^W \frac{\partial f}{\partial I_{c,i,j}}(I_{c,i,j} - I_{0 c, i, j}) \quad  being \quad f(I_0) = 0
\label{eq:relevance-approximation}
\end{equation}

Equation \ref{eq:relevance-approximation} is per se an explanation of $f(I)$ dependent only of the derivative and of $I_0$. The main problem of this approach is finding a valid root that is close under the euclidean norm to the analyzed image $I$. We are approximating the function with a order 1 Taylor expansion and the residuum is proportional to the euclidean distance between both points. Different ways for finding $I_0$ have been proposed. For example, doing a unsupervised search of $f(I)$ over the training set looking for those images reporting $f(I)$ near 0 and averaging them for finding $I_0$.

\subsection{Deep Taylor decomposition}

Deep Taylor decomposition \cite{montavon2017explaining} uses approximation that combines the layer-wise and the Taylor type models. Being compositional the nature of deep learning models, this approach supposes also the decomposability of the relevance function, presuming the existence for every node of a partial relevance function $R_i(a_i)$ that depends on the activation. It considers this function unknown and applies a Taylor decomposition through a root point. Summing up all the individual contributions using the relevance conservation property defined in the previous models, makes possible the propagation of the intermediate relevance to eventually reach the input space and come to a heatmap of the total relevance of the prediction. 

\section{Receptive field and Pixel-wise Explanation Model}\label{sec:math}

In this section we describe our contribution to the explanation models. The model is based on the layer-wise relevance propagation model described above. We reformulate one of the properties of the relevance propagation. All the models of the previous section are based on the fact that relevance should be conservative between layers. In our formulation, we consider the score (we rename relevance to score) entering to a node as the combination of two parts: one that can be transformed into a function dependent on the inputs and another one that is constant and that belongs to the own node. The final score continues to be conservative but not through layers. The final score is the sum of the contribution of the studied feature-space (that can be also the pixel space) plus the score contributions of every following layer. The contribution of every following layer depends of the parameters of the layer and in some way of the output activations. The propagated score depends solely on the individual activation inputs of the layer. In such a way, we are able to find a \emph{unique way} for mapping the score of every output to the input space for the network.

The following propositions are assumed:

\begin{proposition}
	The score of every activation in the network is proportional to the activation value:
	\begin{equation}
		 S_k = \lambda_k a_k
	\end{equation}
\end{proposition}

\begin{proposition}
	The score observed as output one layer can be decomposed in two parts, one dependent on the inputs and another one independent from them that is constant: 
	\begin{equation}
		S_o=S_i+S_k
	\end{equation}
	 where: $S_i$ depends on the input activation of that layer and $S_k$ does not depend on the input activation but only on the parameters of the model that is executed in that layer.
\end{proposition}

The propagation model proposed makes a different treatment of the components $S_i$ and $S_k$. On one hand, $S_i$ depends on the input activation arriving from the original image, so during the propagation backwards we separate the second component $S_k$ and we take $S_o^{(l-1)}=S_i^{(l)}$. In the following subsections we explain how to obtain $S_i$ and $S_k$ for different typical block constituents of deep learning networks.

On the other hand, the $S_k$ values obtained from each of the layers are mapped to the input space in a final procedure by means of the corresponding RFs.

\subsection{Score propagation through an activation function node} 

\begin{figure}[h!]
	\centering
	\includegraphics{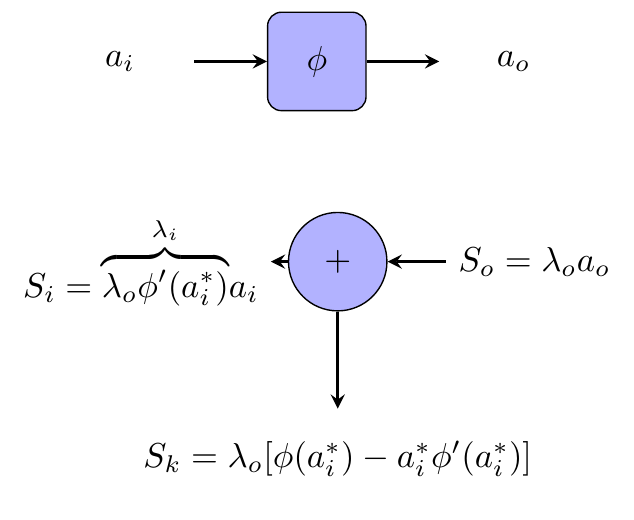}
	\caption{Score propagation through an activation function node}
	\label{fig:score_af}
\end{figure}

In fig. \ref{fig:score_af} we show the activation function node. A input activation $a_i$ is transformed into $a_o = \phi(a_i)$. We know that $S_o = \lambda_o a_o$, substituting $a_o$ we get $S_o = \lambda_o \phi(a_i)$. In order the proposition to be true, we require also that $S_i = \lambda_i a_i$. For ReLU family functions ($\phi(x) = max(0, kx)$), $S_i$ continues verifying the proposition. For other type of activation functions, as we are calculating the score of a particular image, we can consider the network to have parameterizable activation functions. For a particular image we can consider the first order Taylor expansion and see the activation function as a linear function of the form $\phi(a_i) = [\phi(a^*_i) + \phi'(a^*_i)(a_i - a^*_i)]$, where $a^*_i$ is a value close enough to $a_i$ to have a good approximation of $\phi$. After this transformation, the proposition holds for every type of activation function. Substituting and reordering the expression of $S_o$ we obtain that:

\begin{equation}
	S_o = \lambda_o[\phi(a^*_i) - \phi'(a^*_i)a^*_i] + \lambda_o \phi'(a^*_i)a_i
\end{equation}

The score of the output can be splitted in two parts: a constant one that is independent of the activation and belongs to the layer, and another one dependent on the activation. For ReLU, $S_o = S_i$ and $S_k = 0$.

\subsection{Score propagation through a batch normalization node} 

The function implemented in a batch normalization node is $a_o = \beta + \gamma (\frac{a_i - \mu}{\sigma})$. Having $S_o = \lambda_o a_o$, $S_o$ is also $S_o = \lambda_o ( \beta + \gamma (\frac{a_i - \mu}{\sigma}))$. Reordering the expression, we can separate the input independent constants: 

\begin{equation}
	S_o = \lambda_o (\beta - \gamma \frac{\mu}{\sigma}) + \lambda_o \frac{\gamma}{\sigma}a_i
\end{equation}

\begin{figure}[h!]
	\centering
	\includegraphics{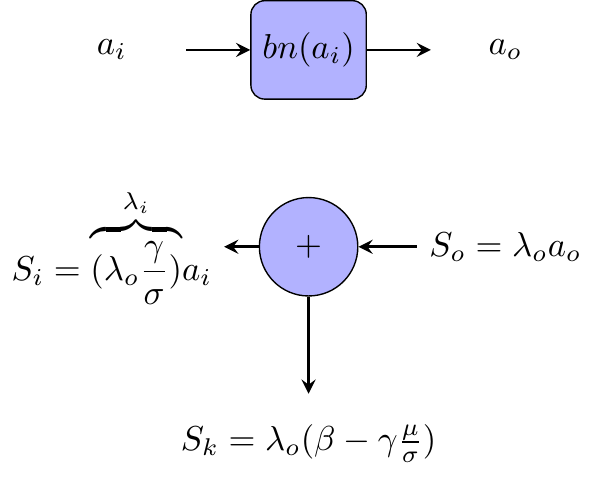}
	\caption{Score propagation through an batch normalization node}
	\label{fig:score_bn}
\end{figure}

As we see, the output score can be exactly splitted into a constant value $S_k = \lambda_o (\beta - \gamma \frac{\mu}{\sigma})$ that is a inherent property of the node and is completely independent of $a_i$ plus $S_i = (\lambda_o \frac{\gamma}{\sigma})a_i = \lambda_i S_i$ that continues to be consistent with the score property proposition, being $\lambda_i = \lambda_o \frac{\gamma}{\sigma}$ (see fig. \ref{fig:score_bn})

\subsection{Score propagation through a convolutional layer}

\begin{figure}[h!]
	\centering
	\begin{subfigure}{0.4\textwidth}
		\includegraphics[scale=0.4]{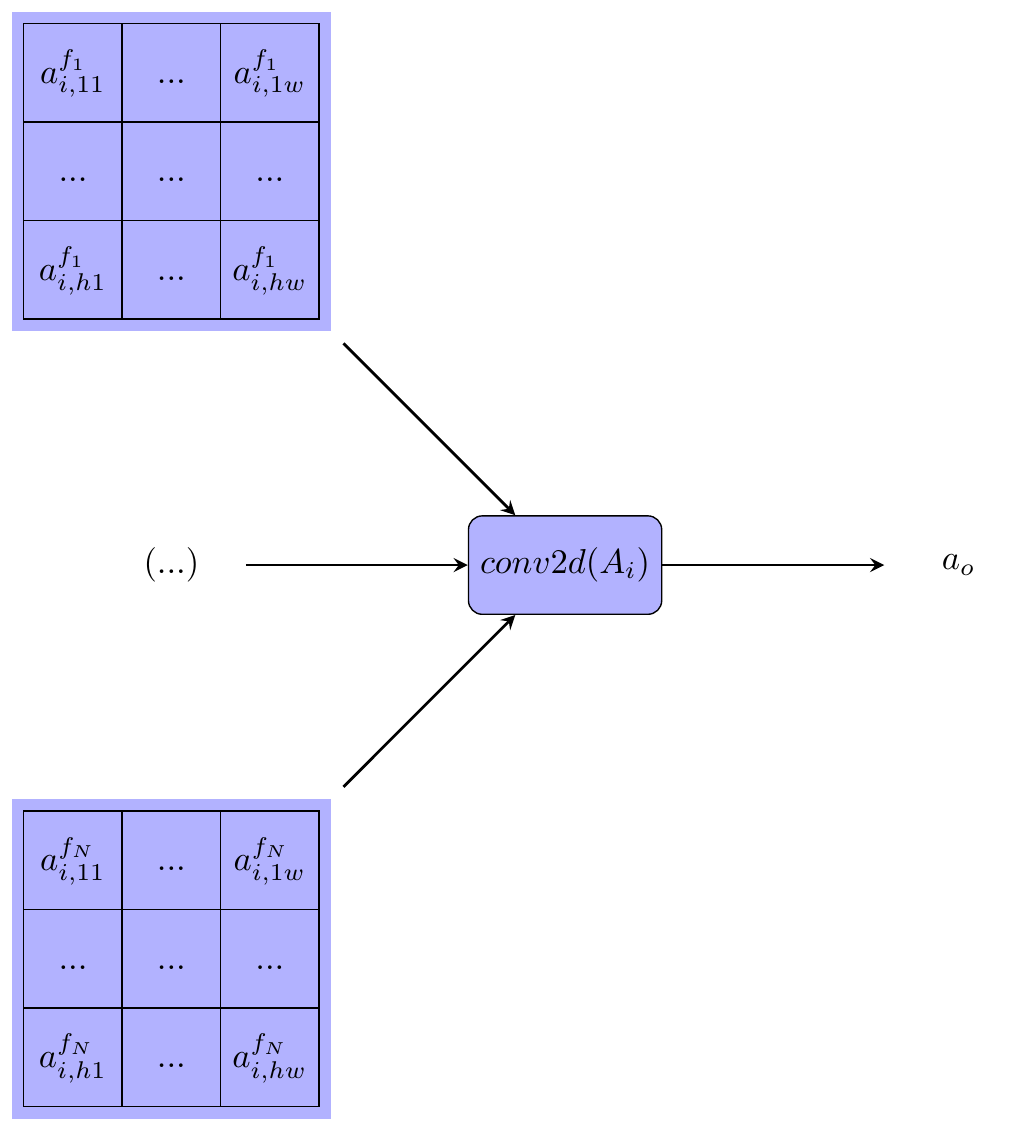}
	\end{subfigure}
	~ 
	\begin{subfigure}{0.4\textwidth}
		\includegraphics[scale=0.4]{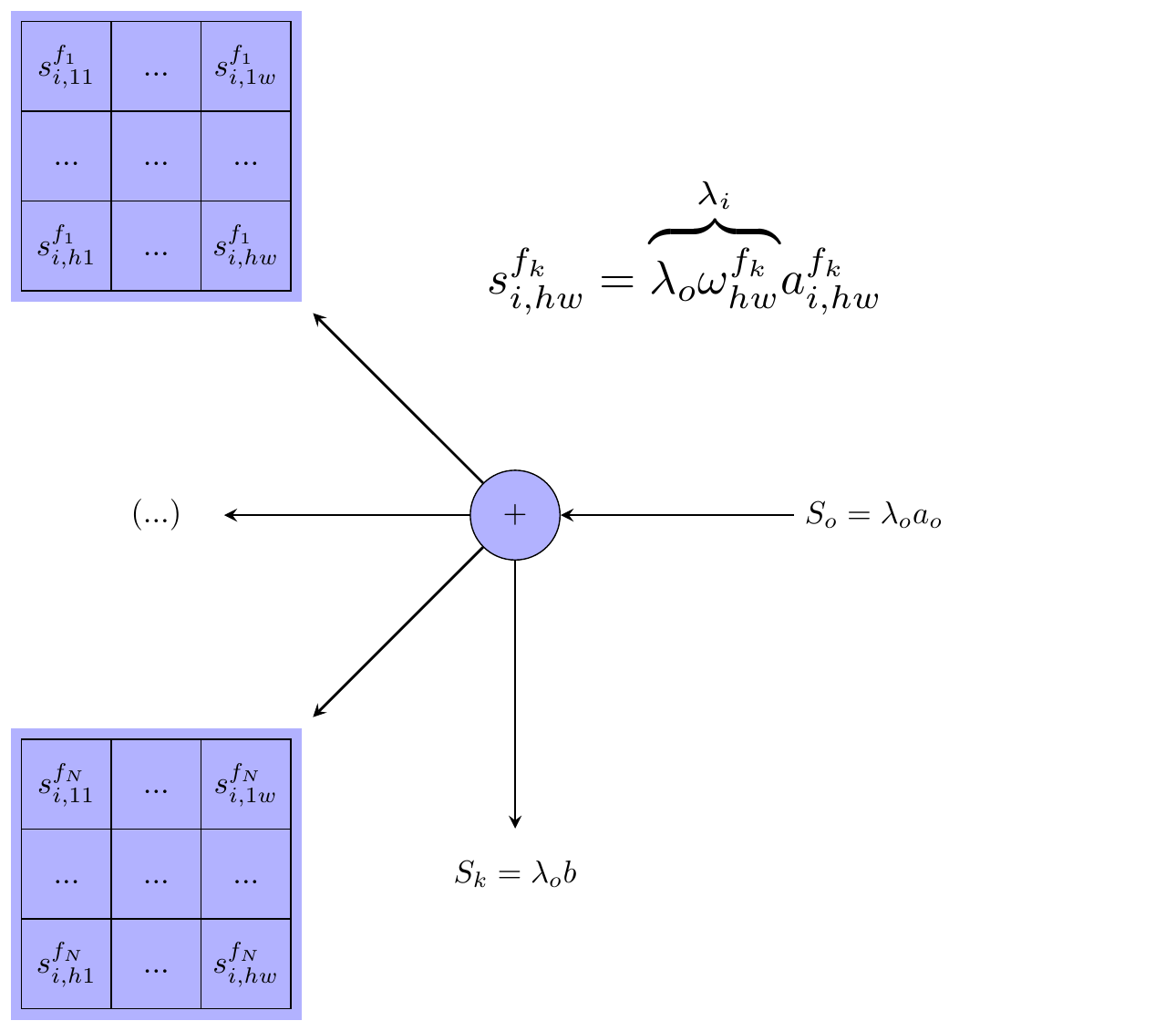}
	\end{subfigure}
	\caption{Convolution score calculation. Score spreads into the different inputs. The bias related part of the score is not backpropagated.}
	\label{fig:convolution_score}
\end{figure}

In the forward propagation of a two dimensional convolution of an image, the set of all the different feature activations of a predefined locality are linearly combined to get the output $a_o$ (see fig. \ref{fig:convolution_score}). Backpropagating a score in a convolutional layer requires to divide it into all its individual components. Every component can be either positive or negative. There is also a bias part, that comes from the inherent nature of the layer and that is not attributable to any of the inputs and that must be treated also as a property of the layer. Due to the nature of the convolution operator, every input node contributes to the calculation of different outputs, that's why every input receives a contribution of the score of different outputs that are summed up.

\subsection{Score propagation through pooling layers}

\begin{figure}[h!]
	\centering
	\begin{subfigure}{0.4\textwidth}
		\includegraphics[scale=0.4]{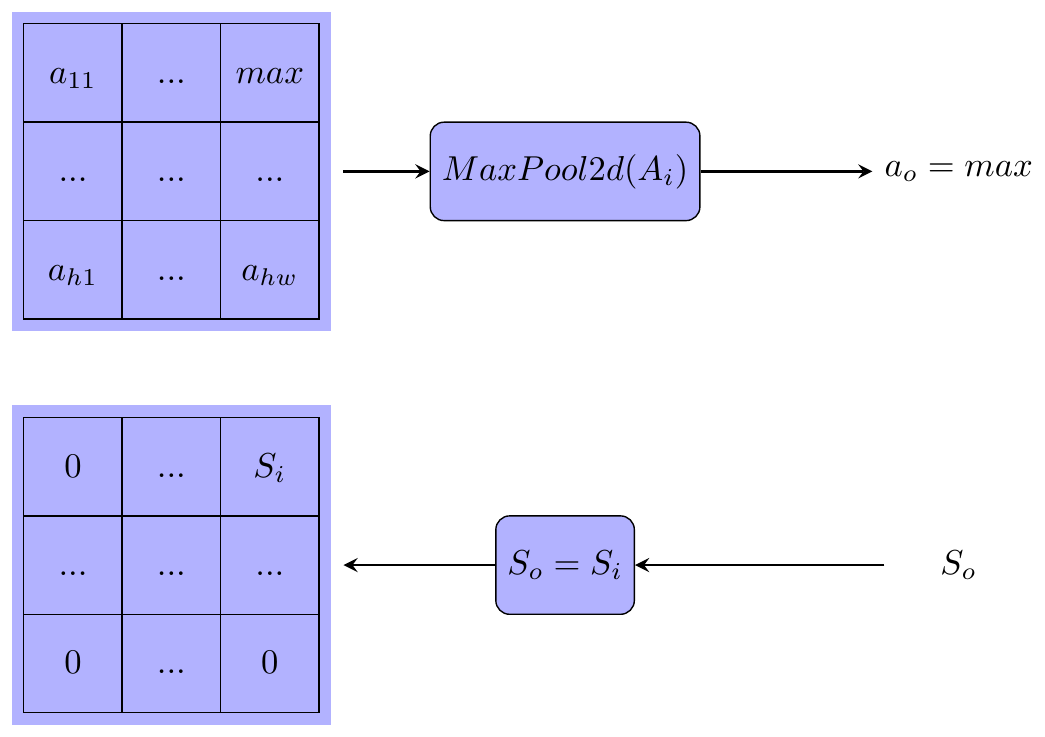}
		\caption{Max-Pooling}
	\end{subfigure}
	~ 
	\begin{subfigure}{0.4\textwidth}
		\includegraphics[scale=0.4]{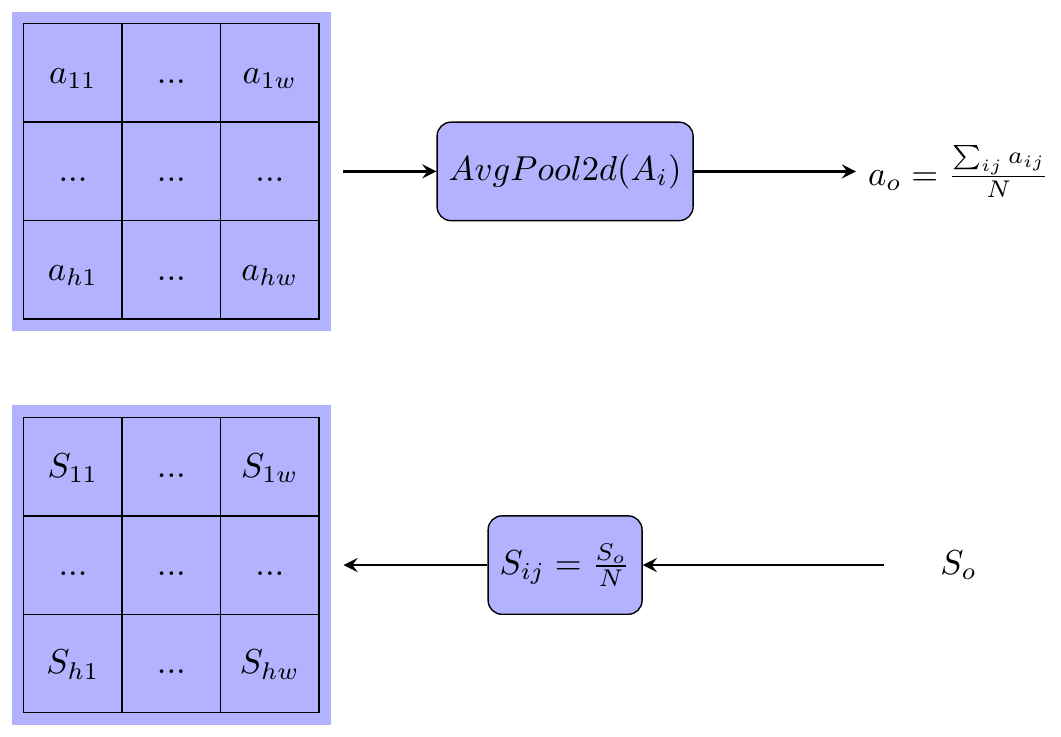}
		\caption{Avg-Pooling}
	\end{subfigure}
	\caption{Score propagation through different pooling layers}
	\label{fig:score_pooling}
\end{figure}

The score propagation through a max-pooling layer is straightforward. For score propagation the value of the score of the output is copied into the input that was selected in the forward pass (see fig. \ref{fig:score_pooling}). For average pooling is also straightforward. For score propagation the value of the score is splitted into $N$ equal parts, being $N$ the number of inputs (see fig. \ref{fig:score_pooling}).

\subsection{Score propagation through a fully connected layer} 

\begin{figure}[h!]
	\centering
	\includegraphics{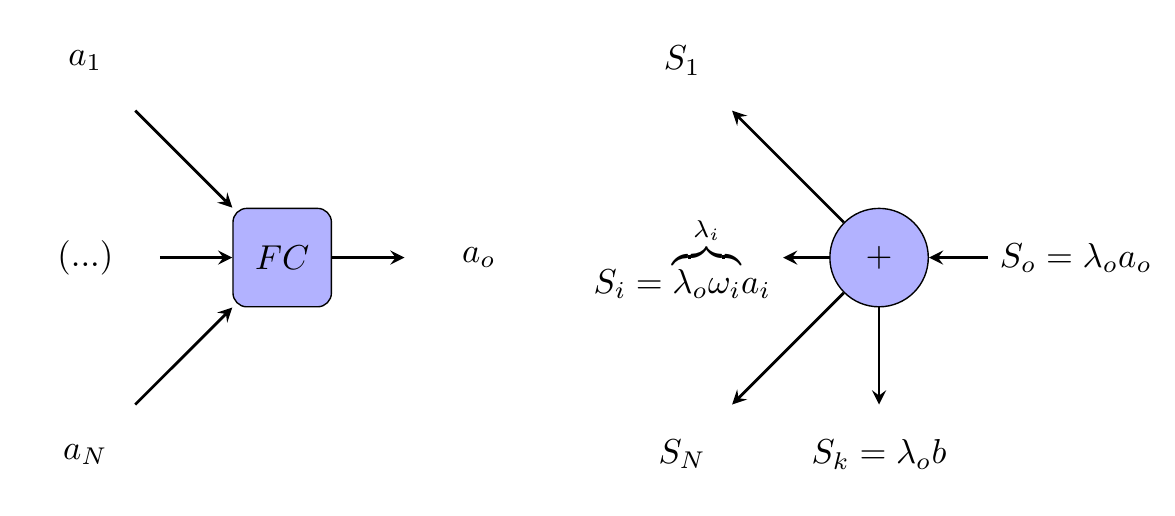}
	\caption{Score propagation through a fully connected node}
	\label{fig:score_fc}
\end{figure}

A fully connected layer is a linear combination of the input activities and the weights. The final score is splitted into the individual elements leaving apart the bias that becomes the score contribution of the own layer (see fig. \ref{fig:score_fc}).

\subsection{Score propagation through a dropout layer}

Dropout in evaluation time acts weighting the output to a value proportional to the dropout probability $a_o = (1-d)a_i$. Inserting this equation into $S_o = \lambda_o a_o$ and applying the conservation of the score through the node ($S_o = S_i$ in this case, due to the absence of constant score) we get that the final equation:

\begin{equation}
 \lambda_i = \lambda_o (1-d)
\end{equation}

\begin{figure}[h!]
	\centering
	\includegraphics{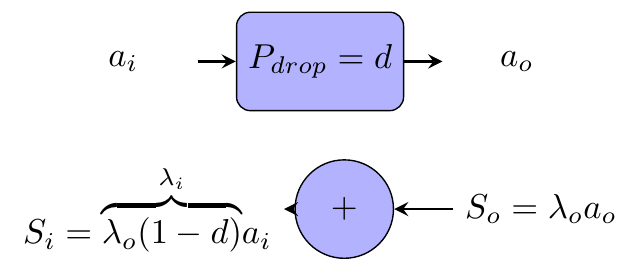}
	\caption{Score propagation through a dropout node}
	\label{fig:score_dropout}
\end{figure}

\subsection{Mapping the score of hidden layers and $S_k$ to input-space}\label{sec:mapping-input}

We have seen that every block has two score constituents: one that is dependent on the inputs and that can be easily forwarded, and another one that depends on the RF, i.e the layer. At this point we are going to transport back also such values to the input-space. From \citep{luo2016understanding} we know that the effective RF is not equal to the theoretical RF. The effective one acts more like a 2D gaussian function where the points located in the borders contribute less than the center ones. Using such property is possible to make an approximate conversion of the hidden-space full and constant scores to the input space using a 2D gaussian prior. For example, for a 20x20 hidden layer with a RF of 189x189 pixels, we know that every of such points is a representation value of a RF of 189x189 in the input space. Having a prior information about the statistical distribution of the input space pixels (in this case gaussian) is possible to go back. Summing up 20x20 gaussian distributions of mean equal to the values of the hidden space and summing up the coincident points is possible to map the distribution in the input space. We fixed $RF = 2\sigma$ as an approximate distribution of the scores, that seems acceptable \citep{luo2016understanding},  98\% of the information of the gaussian is inside the RF. We normalize the function to fit 100\% of the information inside the RF.

\section{Classification Model}\label{sec:class}

\subsection{Data}

In this study we use the EyePACS dataset of the Diabetic Retinopathy Detection competition hosted on the internet Kaggle Platform.  For every patient right and left eye images are reported. All the images are classified by ophthalmologists according to the standard severity scale presented before in \cite{diaclass}. The images are taken in variable conditions: by different cameras, illumination conditions and resolutions. 

The training set contains a total of $75,650$ images; $55,796$ of class 0, $5,259$ of class 1, $11,192$ of class 3, $1,805$ of class 3 and $1,598$ of class 4. The validation set used for hyper-parameter optimization has $3,000$ images; $2,150$ of class 0, $209$ of class 1, $490$ of class 2, $61$ of class 3 and $90$ of class 4. The test set, used only one time for generalization evaluation, contains a total of $10,000$ images; $7,363$ of class 0, $731$ of class 1, $1,461$ of class 2, $220$ of class 3 and $225$ of class 4. 

This dataset is not so rich and well tagged as the used in \citep{doi:10.1001/jama.2016.17216} but allows to train models near human expertise that are useful to show the purposes of our work, that is not only a good performance of the results but mainly study the pixel interpretability of the conclusions (final classification) given by the model.

\subsection{Prediction model}

The model calculates $P(\mathcal{C} | \mathcal{I})$ using as a last layer a $SoftMax$ function over the values after the last linear combinations of the features. This probability is calculated as $P(\mathcal{C} | \mathcal{I}) = \frac{\me^{S_{i}}}{\sum_{j=1}^{C} \me^{S_{j}}}$. Let's call $S_{C}$ the score of the class C, being $S_C$ the final value of each output neuron before applying the $Softmax$. $SoftMax$ function is required for calculating the probability of every class, but in case of being interested only on $argmax(Softmax)$, we needn't evaluate $Softmax$ because $argmax(S_i) = argmax(softmax(S_i))$.

Deep neural network model design up to know is driven mainly by experience. Nowadays there is still more an art than a science and lacks a systematic way for designing the best architecture for solving a problem. In previous works (see \cite{jdelatorre-2016} and \cite{jdelatorre-2017}) we have tested different kinds of architectures that allow us to have a previous knowledge of which kind of models work better for solving this particular classification task.

Using the previous experience in such works we summarize a set of guidelines that ruled the final model selection. These design principles applicable to this the DR particular application, and that are explained below, are: use an optimal image resolution, use all the image information available, use a fully convolutional neural network, use small convolutions, adapt the combination of convolution sizes and number of layers to have a final RF as similar as possible to the image size, use ReLU as activation function, use batch normalization in every layer, use QWK as a loss function, use a efficient number of features and use a linear classifier as the last layer.

\paragraph{Use an optimal image resolution} On one hand the size of the input image has a great importance in the classification results. In this problem in other papers like \cite{jdelatorre-2016} is shown that better results can be achieved with retine diameters of 512 pixels than the ones obtained with 384, 256 or 128 pixels. Some tests done using greater densities than 512 pixel/diameter seem to not improve significantly the classification rates. On the other hand, the hardware of the calculation devices fix a limitation on the available resources. Input image size has a great impact on the memory and calculation time required for the training and test of the deep neural network models. In this work we tested models of 128, 256, 384, 512, 640, 724, 768 and 892 pixels of retine diameter. With this dataset, diameters greater than 640 does not seem to report better results. The optimal size and the used in this study is a retine diameter equal to 640 pixels.

\begin{figure}[h]
	\centering
	\begin{subfigure}[b]{.45\textwidth}
			\centering
			\resizebox{1.0\textwidth}{!}{
				\includegraphics[width=1.0\textwidth]{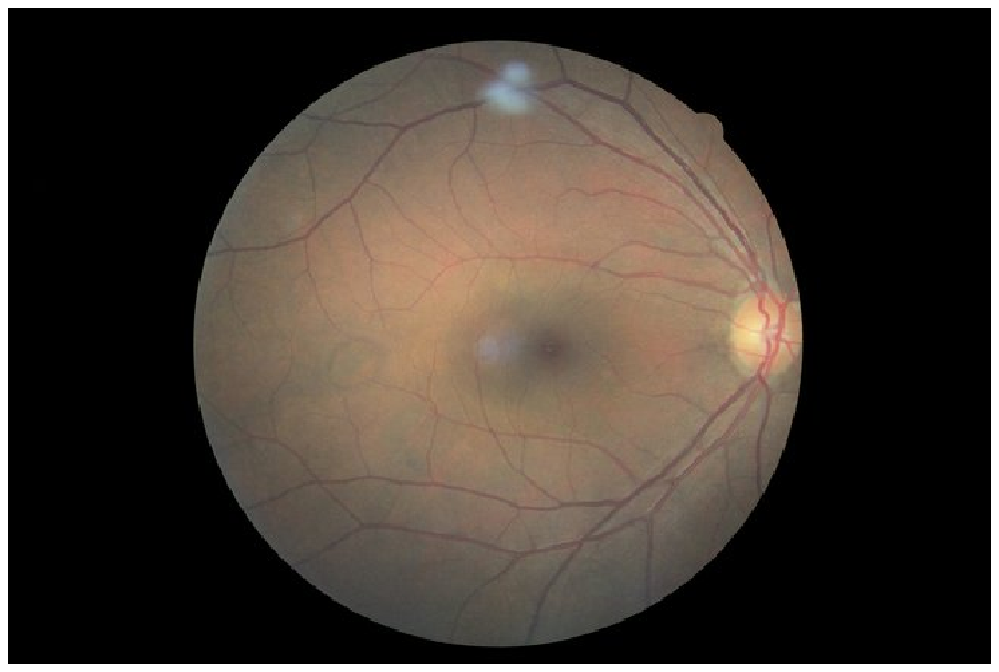}
			}
			\caption{Original 4752x3168 pixels}	
	\end{subfigure}
	\begin{subfigure}[b]{.40\textwidth}
			\centering
			\resizebox{.74\textwidth}{!}{
				\includegraphics[width=1.0\textwidth]{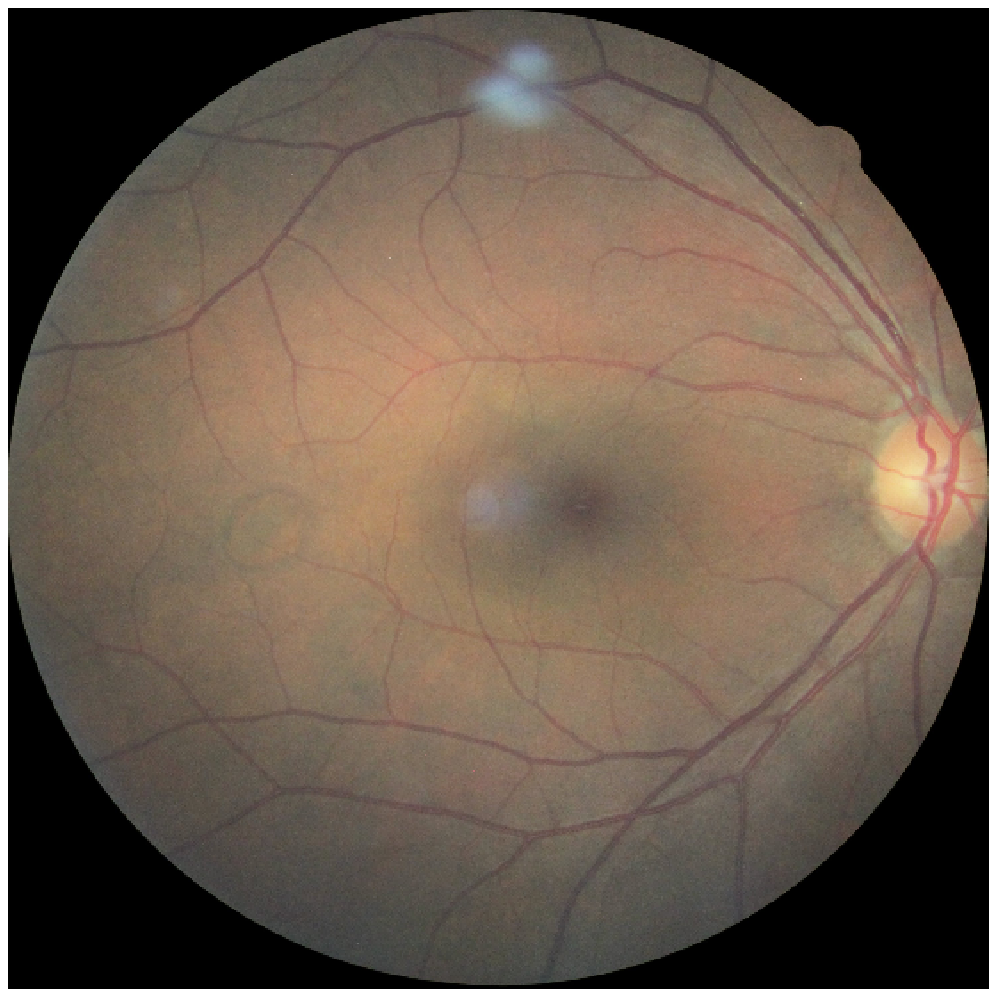}
			}
			\caption{Trim\&resize 640x640 pixels}
	\end{subfigure}
	\caption{A training sample showing the preprocessing treatment}
	\label{fig:preprocessing}
\end{figure}

\paragraph{Use all the available image information} In previous studies published in \citep{jdelatorre-2016} due to hardware limitations the classification models were designed using limited input information, using only part of the available input, requiring ensembling solutions to combine the results from evaluating different parts of the same retine. A 512x512 input image model was used with a random selection of a rotated square (diagonal equal to the retine diameter). In this way only a 64\% of the retine information available was used in the classification prediction. On test time five rotated versions of the input where averaged in order to get a better evaluation result. In this paper we use a network that receives all the input information available not requiring ensembling on test time. Only background located further from the diameter is removed (see fig \ref{fig:preprocessing}).

\paragraph{Use a fully convolutional neural network} Convolutional neural networks (CNN) are computationally more efficient than fully connected ones. CNNs are ideal for exploiting the typical high local pixel correlations present in images.

\paragraph{Use small size convolutions} The stacking of small size convolutions is more efficient than the usage of big size convolutions. With a lower number of parameters is possible to generate more nonlinear relationships between the pixels that only using a unique convolution of higher size. Following this philosophy only 3x3 convolutions have been used in the feature layers. 

\begin{wrapfigure}{r}{0.30\textwidth}
	\centering
	\includegraphics[width=0.30\textwidth]{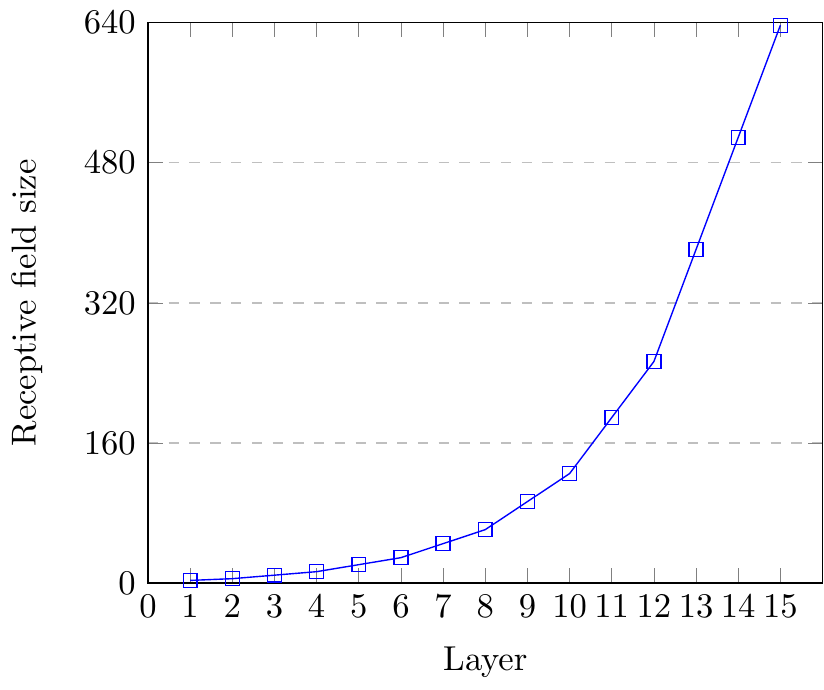}
	\caption{Model RF size growth}
	\label{fig:rf_graph}
\end{wrapfigure}

\paragraph{Adapt convolution sizes and number of layers to get a RF as similar as possible to the image size} One important aspect of CNNs is the RF size. RF defines the theoretical space covered by a convolution in the input space. The ideal case is having a RF in the last layer equal to the image size, because in such a way we are sure that all the information available is used. RFs greater than image size are inefficient, that's why sometimes can be necessary to slightly modify the convolution sizes of some layer to get the desired one. Figure \ref{fig:rf_graph} shows the RF growth of our model.

\paragraph{Use rectified linear unit (ReLU) as activation function} ReLU is a computationally efficient activation function that is very suitable to be used with very deep convolutional neural networks\citep{Dahl2013}. Derivatives and Scores as we will see are easily calculated. We have tested other activation functions such as LeakyReLU, ELU and SeLU reporting similar and even worse results, introducing complexity to the model without a clear advantage in the final result.

\paragraph{Use batch normalization in every layer} Batch normalization \citep{batch-norm} stabilize the training and accelerates convergence. In this problem there is a great difference between using batch normalization or not, to the point that not using it makes very difficult or even impossible the training.

\paragraph{Use QWK as a loss function} For multi-class classification the standardized loss function to use is the logarithmic loss \citep{Goodfellow-et-al-2016}. In \citep{DELATORRE2017} is shown that for ordinal regression problems, where not only a multi-class classification is taking place but also there is possible to establish a sorting of the classes based on some hidden underlying causes, QWK-loss can also be used with better results. The properties of this function as a loss function have been widely studied in \cite{jdelatorre-2017}. The difference in the performance of the final results is very high. Optimizing directly QWK allows getting better classification results.

\paragraph{Use a linear classifier as a last layer} For simplicity and interpretability of the model we expect the model to disentangle completely the features required for the classification. The final classification is required to be a linear combination of the features of the last layer.

\begin{wrapfigure}{r}{0.30\textwidth}
	\centering
	\includegraphics[width=0.30\textwidth]{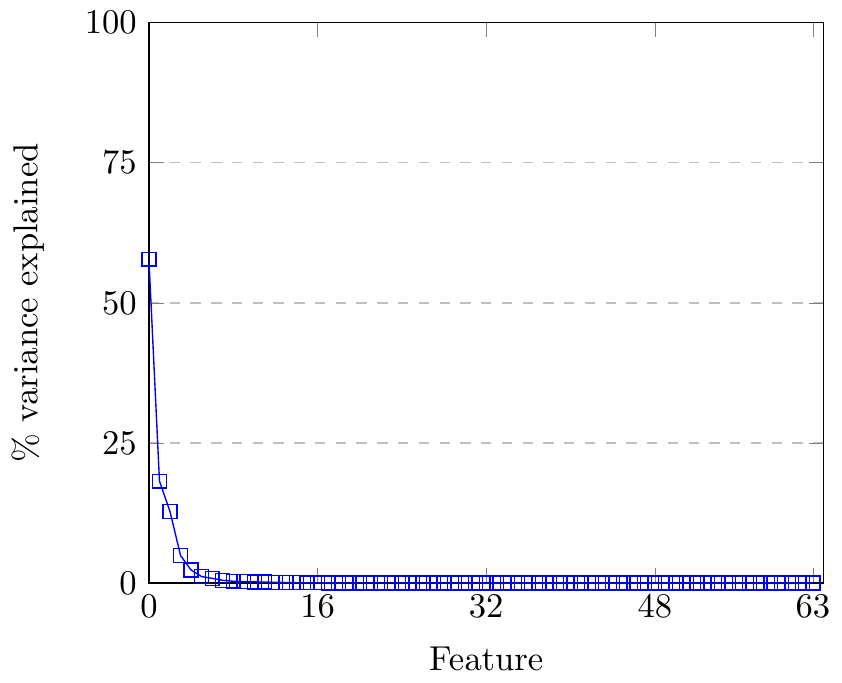}
	\caption{Feature space cummulative PCA variance computed over the training set}
	\label{fig:pca_graph}
\end{wrapfigure}

\paragraph{Use a efficient number of features} With infinitely number of resources we can use a big network. In our case we have limited resources and not only this but we would like also to be able to implement the result in devices with low resources. In this way we have tested networks of different sizes and in order to check the redundancy of the information, we made a principal component analysis (PCA) in the feature space of the last layer, arriving to the conclusion that about 32 of the features explain 98.3 \% and 48 features, 99.997\% of the total variance. We studied different configurations using different number of features from 512 to 32. Values of 32 showed a reduction in performance that increased when increasing the features to 64. Higher number of features did not improve the results. In figure \ref{fig:pca_graph} we show the variance explained by final feature vector space.

\begin{figure}[h!]
	\begin{subfigure}[b]{.30\textwidth}
		\centering
		\includegraphics[width=\textwidth]{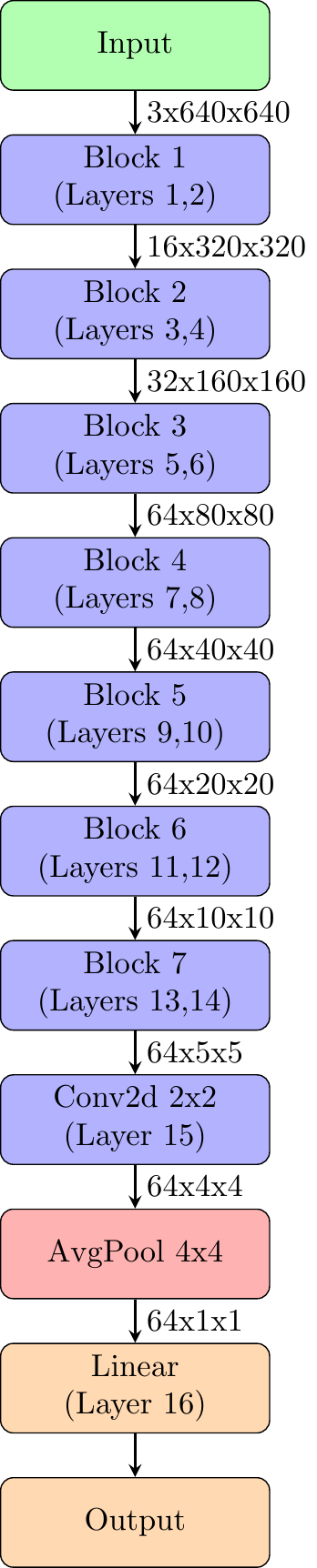}
	\end{subfigure}
	\hfill    
	\begin{subfigure}[b]{.30\textwidth}
		\centering
		\includegraphics[width=\textwidth]{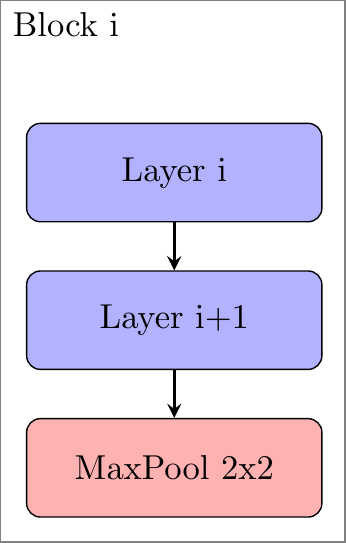}\\
		\vspace{2cm}
		\includegraphics[width=\textwidth]{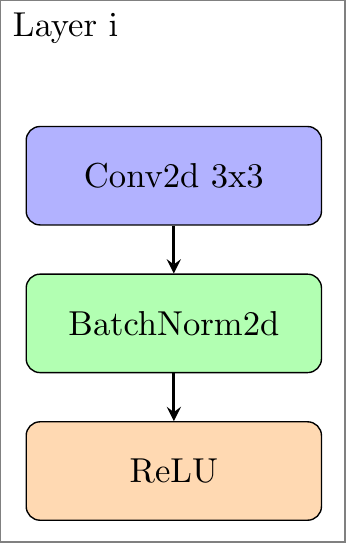}\\
		\vspace{2cm}
	\end{subfigure}
	\hfill 
	\caption{Classification model for DR prediction}  
	\label{fig:drmodel} 
\end{figure}

\paragraph{Model description} The model use a 3x640x640 input image obtained from a minimal preprocessing step where only the external background borders are trimmed and later resized to the required input size (see fig. \ref{fig:preprocessing}). Figure \ref{fig:drmodel} shows a block diagram of the model. It is a CNN of 391,325 parameters, divided in 17 layers. Layers are divided into two groups: the feature extractor and the classifier. The feature extraction has 7 blocks of 2 layers. Every layer is a stack of a 3x3 convolution with stride 1x1 and padding 1x1 followed by a batch normalization and a ReLU activation function. Between every block a 2x2 max-pooling operation of stride 2x2 is applied. After the 7 blocks of feature extraction, the RF of the network has grown till reaching 637x637, that is approximately equal to the input size 640x640 (see fig \ref{fig:rf_graph} to see the RF of every layer). Afterwards, the classification phase takes place using a 2x2 convolution. A 4x4 average-pooling reduces the dimensionality to get a final 64 feature vector that are linearly combined to obtain the output scores of every class. A soft-max function allows the conversion of the scores to probabilities to feed the values to the proper cost function during the optimization process. The feature extractor has 16 filters in the first block, 32 in the second and 64 in all the other.

\subsection{Training Procedure}

The training set has 75,650 images and the validation set used for hyper-parameter selection 3,000. Notice that the image set is highly imbalanced. In order facilitate the learning, the training set is artificially equalized using data augmentation techniques \citep{Krizhevsky:2012} based on $0-180^{\circ}$ random rotation, X and Y mirroring and contrast and brightness random sampling.

A random initialization based in the Kaiming\&He approach \citep{kaiming} is used for all the networks. All models are optimized using a batch based first order optimization algorithm called Adam \citep{DBLP:journals/corr/KingmaB14}. The loss function used for optimizing the model is the qwk-loss, with a batch size of $15$ and a learning rate of $3x10^{-4}$ \citep{DELATORRE2017}. 

For every batch, the images are chosen randomly from the training set, with repetition. Data augmentation techniques are applied to increase the diversity of the classes (random rotations and brightness and contrast modifications). 

After training the network, a linear classifier formed by the combination of the 128 features of the two eyes of the patient is trained. In this way is possible to increase further the prediction performance of the model using all the information available of the patient.

\section{Results}\label{sec:results}

\subsection{Classification}

The model is trained for 300 epochs,reaching a QWK evaluation metric over the validation set of $0.814$. The value achieved in the never seen before test set is of $0.801$. Using a linear classifier for combining the features of both eyes $QWK_{test}$ reaches $0.844$. Expert ophthalmologist report QWK inter-rating agreement values in the $0.80s$. Training the model as a multi-class classification model facilitates the encoding of the required features for distinguishing between the different severity levels of the disease. Training the model for an aggregated detection (grouping the positive classes) will for sure increment the accuracy, at the prize of missing the coding of important features that separate positive classes (1 to 4). In our case we want the model to learn such a differences to visualize them in the explanation model, that's why in our case is better to use all the information available about the gradation of disease (intermediate classes) in order to force the model to encode the required features for separating between the intermediate classes at the prize obviously of reducing accuracy in the correct predictions. In this way after back-propagating the explanations we could get the scores that the model gives in the evaluation of the different classes for the same image, allowing the expert to include its own expertise in the final decision.

\subsection{Explanations}

In this subsection we describe the steps followed in the score calculation of a test set sample. After that we present also a set of test images with its calculated score maps for the predicted class. 

\begin{wrapfigure}{r}{0.3\textwidth}
	\centering
	\includegraphics[width=0.20\textwidth]{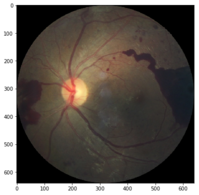}
	\caption{Retine test sample}
	\label{fig:retine_test1}
\end{wrapfigure}

The image shown in fig. \ref{fig:retine_test1}, is tagged in the test set as class 4. After feeding it into the model we get the next classification scores (previous to soft-max): $C_0 = -638.9$, $C_1 = -379.7$, $C_2 = -114.6$, $C_3 = +62.8$ and $C_4 = +167.1$. Being $C_4$ the highest value, the image is correctly classified as class 4. 

\begin{figure}[h!]
	\centering
	\includegraphics[width=\textwidth]{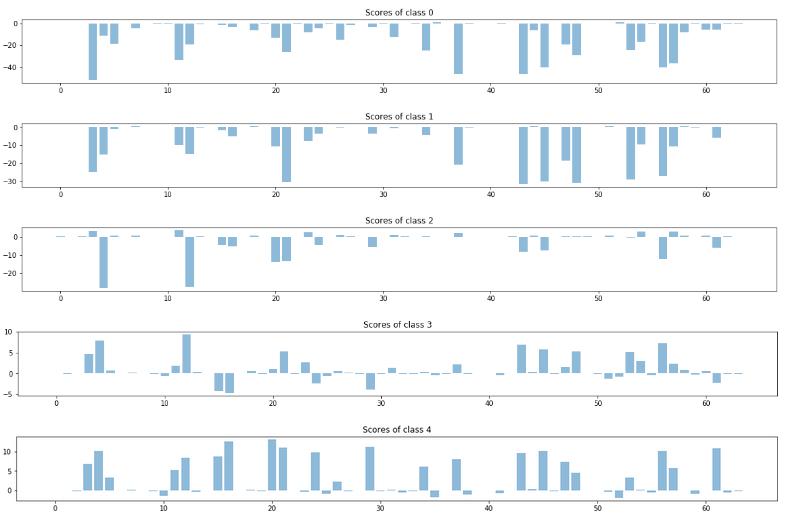}
	\caption{Layer 16 feature scores for considered test sample}
	\label{fig:feature_scores}
\end{figure}

Fig. \ref{fig:feature_scores} shows layer 16 individual feature scores. It can be observed in the bar plot how the same features score different for the prediction of the different classes.

For visualization purposes, layer scores are presented considering the layer as a unique block combination of \emph{convolution - batch normalization - ReLU}. The output of this function block can be mathematically expressed as:

\begin{equation}
O = max(0, \beta + \gamma(\frac{WI + b - \mu}{sigma})
\end{equation}

being the output score:

\begin{equation}
S_O = \lambda (\beta + \gamma \frac{b - \mu}{\sigma}) + \lambda \frac{\gamma}{\sigma}WI
\end{equation}

In this way the output score can be splitted in the two parts: the score input, $S_I = \lambda \frac{\gamma}{\sigma}WI$ and the constant score of the layer, $S_k = \lambda (\beta + \gamma \frac{b - \mu}{\sigma})$. 

\begin{figure}[h!]
	\centering
	\begin{subfigure}[b]{\textwidth}
		\includegraphics[width=\textwidth]{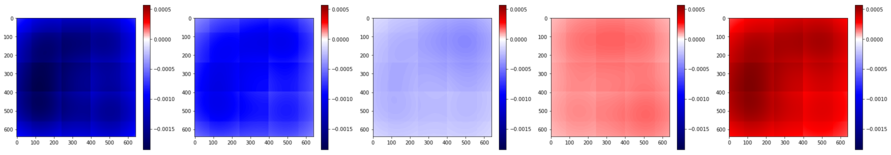}
		\caption{$S^{(15)}$, RF=637x637}
		\label{fig:score_rf637}
	\end{subfigure}
	
	\begin{subfigure}[b]{\textwidth}
		\includegraphics[width=\textwidth]{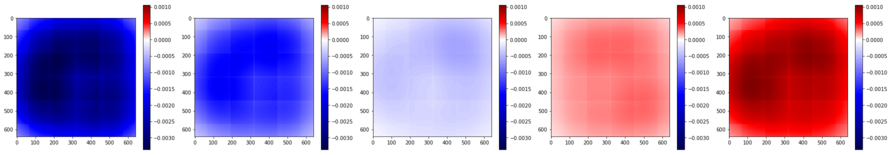}
		\caption{$S^{(14)}$, RF=509x509}
		\label{fig:score_rf509}
	\end{subfigure}
	
	\begin{subfigure}[b]{\textwidth}
		\includegraphics[width=\textwidth]{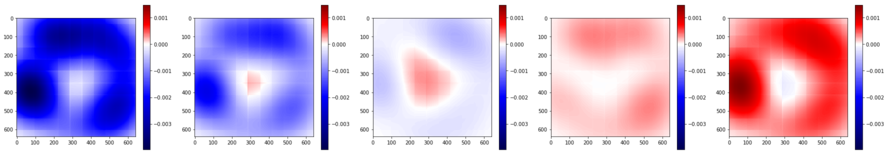}
		\caption{$S^{(13)}$, RF=381x381}
		\label{fig:score_rf381}
	\end{subfigure}
	
	\begin{subfigure}[b]{\textwidth}
		\includegraphics[width=\textwidth]{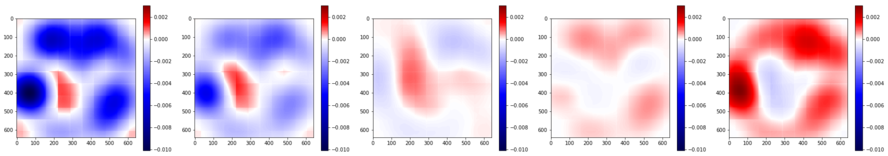}
		\caption{$S^{(12)}$, RF=253x253}
		\label{fig:score_rf253}
	\end{subfigure}
	
	\begin{subfigure}[b]{\textwidth}
		\includegraphics[width=\textwidth]{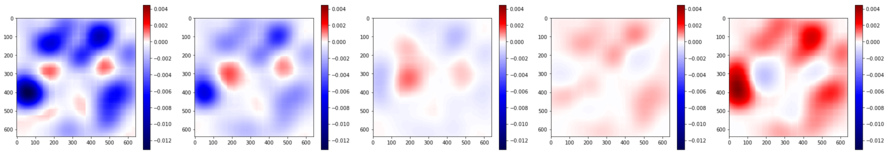}
		\caption{$S^{(11)}$, RF=189x189}
		\label{fig:score_rf189}
	\end{subfigure}
	
	\begin{subfigure}[b]{\textwidth}
		\includegraphics[width=\textwidth]{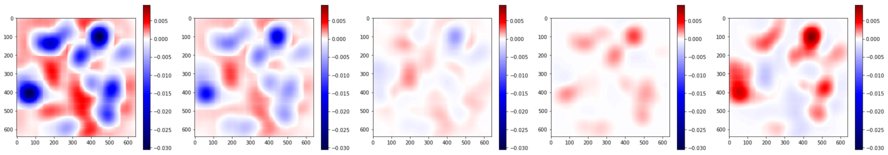}
		\caption{$S^{(10)}$, RF=125x125}
		\label{fig:score_rf125}
	\end{subfigure}
	
	\caption{Full explanation of the classification of test retine image (layers 15-10). From left to right aggregated score maps for class 0 to class 4 of every referred layer}
	\label{fig:test1_score_explanation_rf}
\end{figure}

\begin{figure}[h!]
	\centering
	\begin{subfigure}[b]{\textwidth}
		\includegraphics[width=\textwidth]{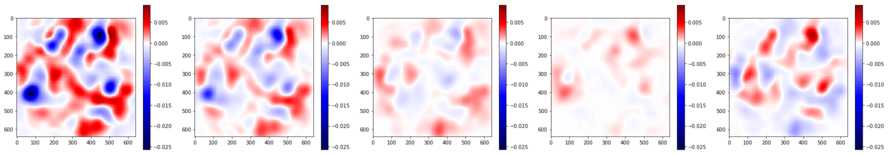}
		\caption{$S^{(9)}$, RF=93x93}
		\label{fig:score_rf93}
	\end{subfigure}
	
	\begin{subfigure}[b]{\textwidth}
		\includegraphics[width=\textwidth]{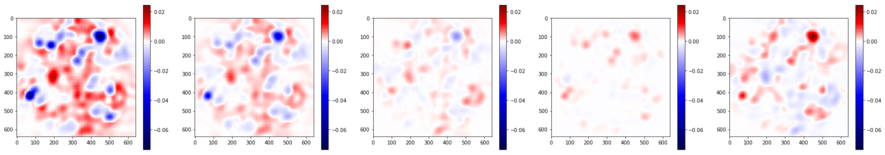}
		\caption{$S^{(8)}$, RF=61x61}
		\label{fig:score_rf61}
	\end{subfigure}
	
	\begin{subfigure}[b]{\textwidth}
		\includegraphics[width=\textwidth]{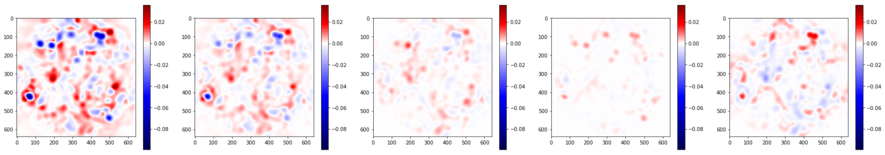}
		\caption{$S^{(7)}$, RF=45x45}
		\label{fig:score_rf45}
	\end{subfigure}
	
	\begin{subfigure}[b]{\textwidth}
		\includegraphics[width=\textwidth]{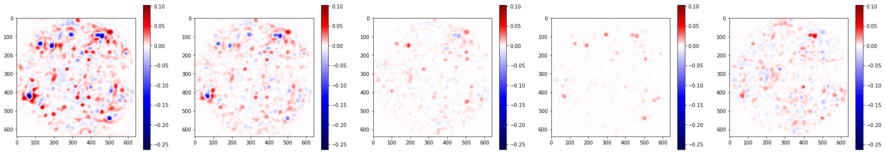}
		\caption{$S^{(6)}$, RF=29x29}
		\label{fig:score_rf29}
	\end{subfigure}
	
	\begin{subfigure}[b]{\textwidth}
		\includegraphics[width=\textwidth]{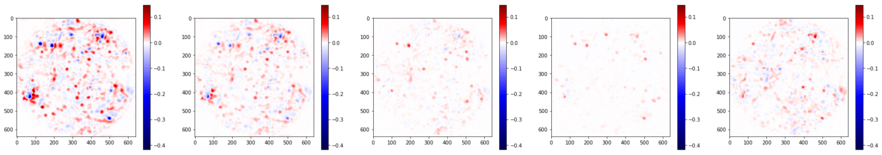}
		\caption{$S^{(5)}$, RF=21x21}
		\label{fig:score_rf21}
	\end{subfigure}
	
	\begin{subfigure}[b]{\textwidth}
		\includegraphics[width=\textwidth]{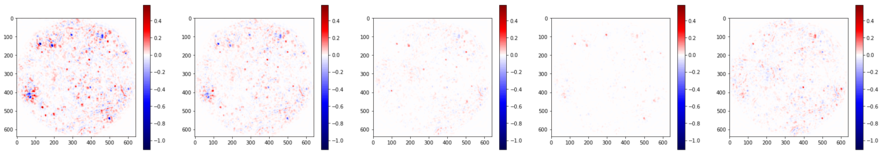}
		\caption{$S^{(4)}$, RF=13x13}
		\label{fig:score_rf13}
	\end{subfigure}
	
	\caption{Full explanation of the classification of test retine image (layers 9-4). From left to right aggregated score maps for class 0 to class 4 of every referred layer}
	\label{fig:test1_score_explanation2_rf}
\end{figure}

\begin{figure}[h!]
	\centering
	\begin{subfigure}[b]{\textwidth}
		\includegraphics[width=\textwidth]{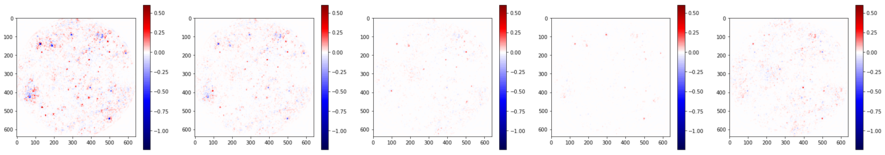}
		\caption{$S^{(3)}$, RF=9x9}
		\label{fig:score_rf9}
	\end{subfigure}

	\begin{subfigure}[b]{\textwidth}
		\includegraphics[width=\textwidth]{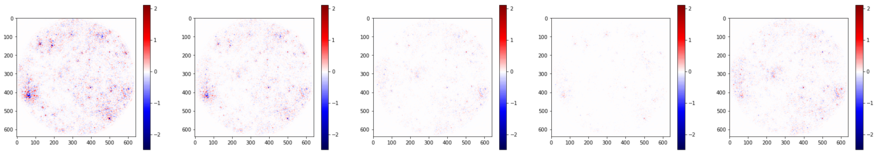}
		\caption{$S^{(2)}$, RF=5x5}
		\label{fig:score_rf5}
	\end{subfigure}
	
	\begin{subfigure}[b]{\textwidth}
		\includegraphics[width=\textwidth]{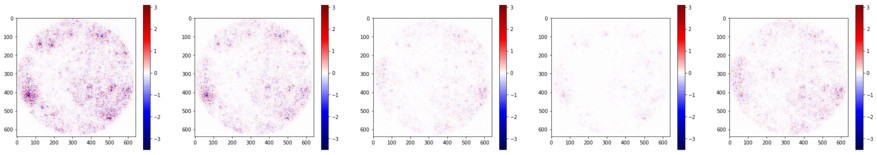}
		\caption{$S^{(1)}$, RF=3x3}
		\label{fig:score_rf3}
	\end{subfigure}
	
	\begin{subfigure}[b]{\textwidth}
		\includegraphics[width=\textwidth]{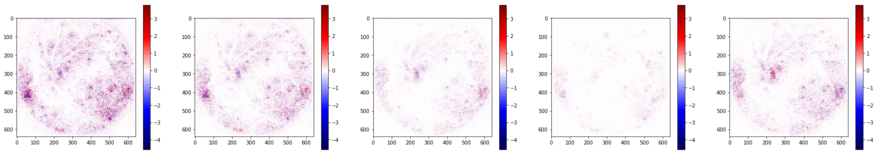}
		\caption{$S^{(input)}$}
		\label{fig:score_input}
	\end{subfigure}

	\caption{Full explanation of the classification of test retine image (layer 3-0). From left to right aggregated score maps for class 0 to class 4 of every referred layer}
	\label{fig:test1_score_explanation3_rf}
\end{figure}

\begin{figure}[h!]
	\centering
	\begin{subfigure}[b]{\textwidth}
		\includegraphics[width=\textwidth]{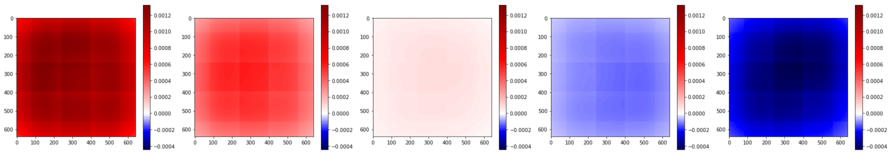}
		\caption{$S_k^{(15)}$, RF=637x637}
		\label{fig:score_k637}
	\end{subfigure}
	
	\begin{subfigure}[b]{\textwidth}
		\includegraphics[width=\textwidth]{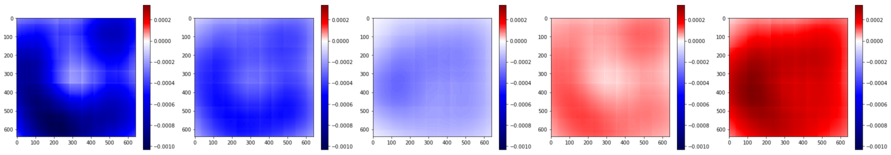}
		\caption{$S_k^{(14)}$, RF=509x509}
		\label{fig:score_k509}
	\end{subfigure}
	
	\begin{subfigure}[b]{\textwidth}
		\includegraphics[width=\textwidth]{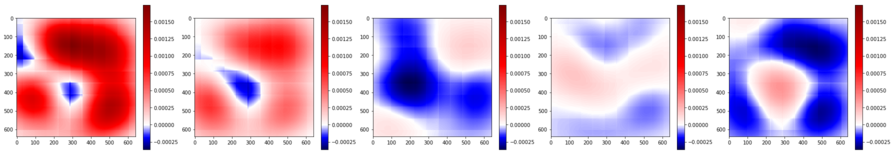}
		\caption{$S_k^{(13)}$, RF=381x381}
		\label{fig:score_k381}
	\end{subfigure}

	\begin{subfigure}[b]{\textwidth}
		\includegraphics[width=\textwidth]{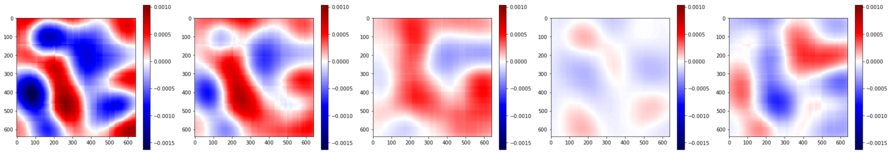}
		\caption{$S_k^{(12)}$, RF=253x253}
		\label{fig:score_k253}
	\end{subfigure}

	\begin{subfigure}[b]{\textwidth}
		\includegraphics[width=\textwidth]{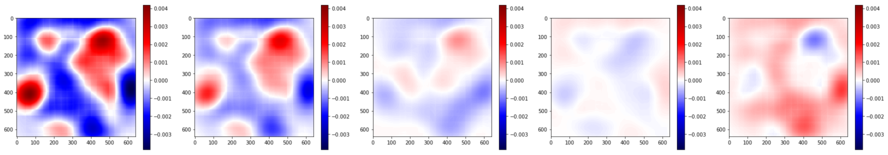}
		\caption{$S_k^{(11)}$, RF=189x189}
		\label{fig:score_k189}
	\end{subfigure}

	\begin{subfigure}[b]{\textwidth}
		\includegraphics[width=\textwidth]{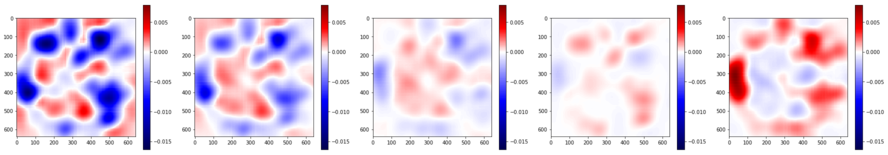}
		\caption{$S_k^{(10)}$, RF=125x125}
		\label{fig:score_k125}
	\end{subfigure}

	\caption{RF dependent constant scores for test sample (layers 15-10). From left to right aggregated score maps for class 0 to class 4 of every referred layer}
	\label{fig:test1_score_explanation_k}
\end{figure}

\begin{figure}[h!]
	\centering
	\begin{subfigure}[b]{\textwidth}
		\includegraphics[width=\textwidth]{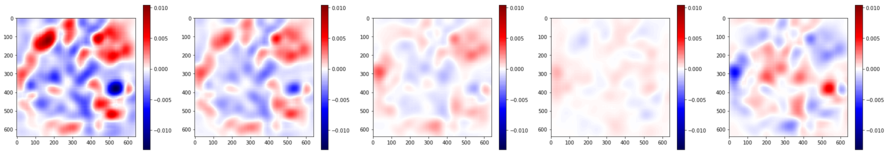}
		\caption{$S_k^{(9)}$, RF=93x93}
		\label{fig:score_k93}
	\end{subfigure}	
	\begin{subfigure}[b]{\textwidth}
		\includegraphics[width=\textwidth]{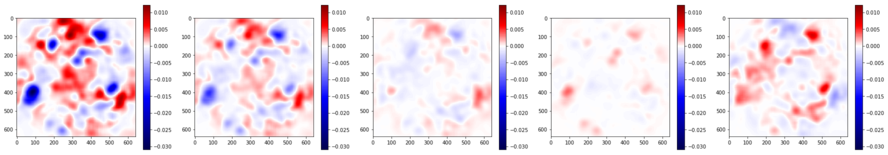}
		\caption{$S_k^{(8)}$, RF=61x61}
		\label{fig:score_k61}
	\end{subfigure}
	
	\begin{subfigure}[b]{\textwidth}
		\includegraphics[width=\textwidth]{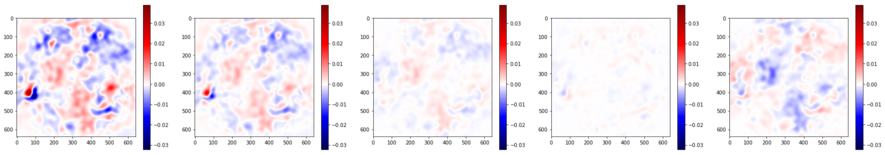}
		\caption{$S_k^{(7)}$, RF=45x45}
		\label{fig:score_k45}
	\end{subfigure}
	
	\begin{subfigure}[b]{\textwidth}
		\includegraphics[width=\textwidth]{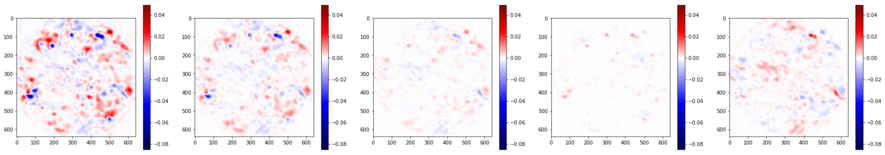}
		\caption{$S_k^{(6)}$, RF=29x29}
		\label{fig:score_k29}
	\end{subfigure}
	
	\begin{subfigure}[b]{\textwidth}
		\includegraphics[width=\textwidth]{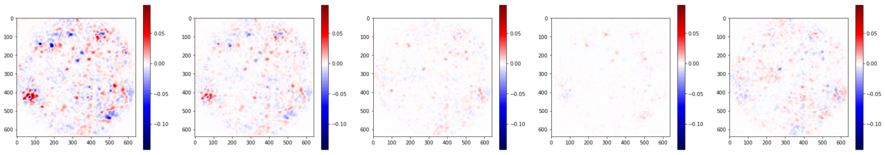}
		\caption{$S_k^{(5)}$, RF=21x21}
		\label{fig:score_k21}
	\end{subfigure}
	
	\begin{subfigure}[b]{\textwidth}
		\includegraphics[width=\textwidth]{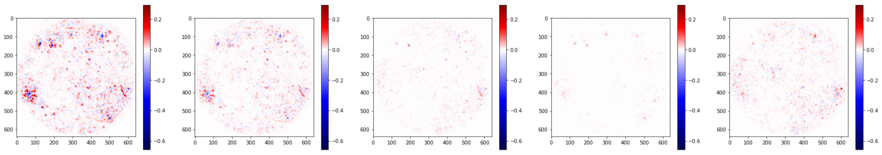}
		\caption{$S_k^{(4)}$, RF=13x13}
		\label{fig:score_k13}
	\end{subfigure}
	
	\caption{RF dependent constant scores for test sample (layers 9-4). From left to right aggregated score maps for class 0 to class 4 of every referred layer}
	\label{fig:test1_score_explanation2_k}
\end{figure}

\begin{figure}[h!]
	\centering
	\begin{subfigure}[b]{\textwidth}
		\includegraphics[width=\textwidth]{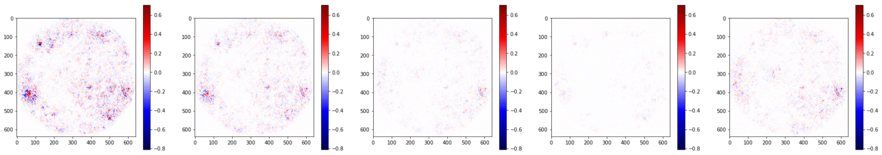}
		\caption{$S_k^{(3)}$, RF=9x9}
		\label{fig:score_k9}
	\end{subfigure}

	\begin{subfigure}[b]{\textwidth}
		\includegraphics[width=\textwidth]{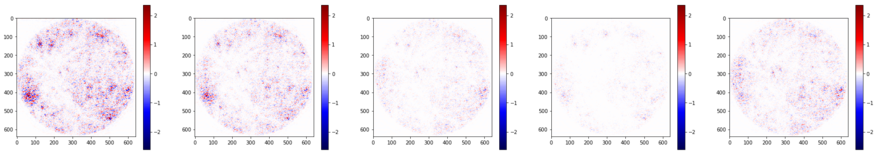}
		\caption{$S_k^{(2)}$, RF=5x5}
		\label{fig:score_k5}
	\end{subfigure}
		
	\begin{subfigure}[b]{\textwidth}
	\includegraphics[width=\textwidth]{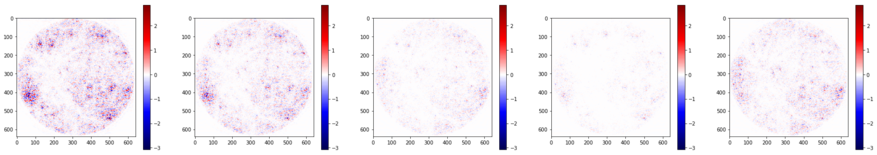}
	\caption{$S_k^{(1)}$, RF=3x3}
	\label{fig:score_k3}
\end{subfigure}

\caption{RF dependent constant scores for test sample (layers 3-1). From left to right aggregated score maps for class 0 to class 4 of every referred layer}
\label{fig:test1_score_explanation3_k}
\end{figure}

Figures \ref{fig:test1_score_explanation_rf}, \ref{fig:test1_score_explanation2_rf}, and \ref{fig:test1_score_explanation3_rf} show the aggregated scores of every hidden layer and of the final output layer. Individual feature scores are first calculated, \emph{receptive field-wise} summed up and mapped into input-space (section \ref{sec:mapping-input}). The same is done for $S_k^{(l)} \quad \forall l \in L$. Figures \ref{fig:test1_score_explanation_k}, \ref{fig:test1_score_explanation2_k} and \ref{fig:test1_score_explanation3_k} show the $S_k$ of all the hidden layers. Fig. \ref{fig:score_input} show the part of score that depend exclusively of the input. 

\begin{figure}[h!]
	\centering
	\begin{subfigure}[b]{0.43\textwidth}
		\includegraphics[width=\textwidth]{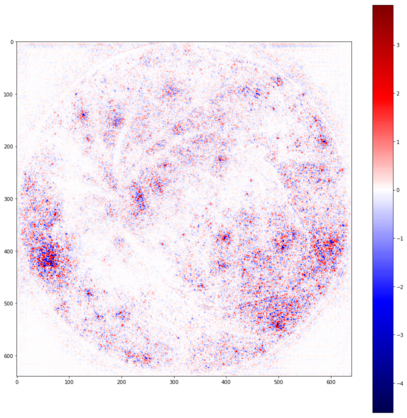}
		\caption{Total score for $C_0$}
	\end{subfigure}~
	\begin{subfigure}[b]{0.43\textwidth}		
		\includegraphics[width=\textwidth]{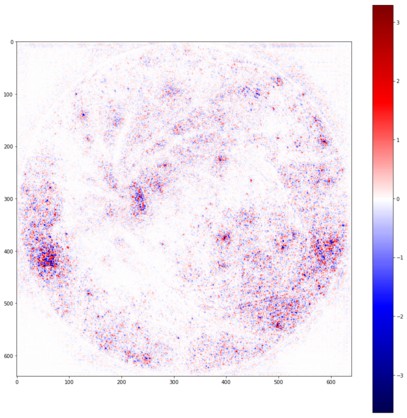}
		\caption{Total score for $C_1$}
		\label{fig:score_total_c1}
	\end{subfigure}
	\begin{subfigure}[b]{0.43\textwidth}
		\includegraphics[width=\textwidth]{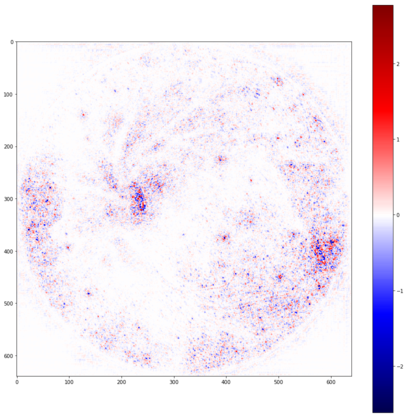}
		\caption{Total score for $C_2$}
		\label{fig:score_total_c2}
	\end{subfigure}~
	\begin{subfigure}[b]{0.43\textwidth}
		\includegraphics[width=\textwidth]{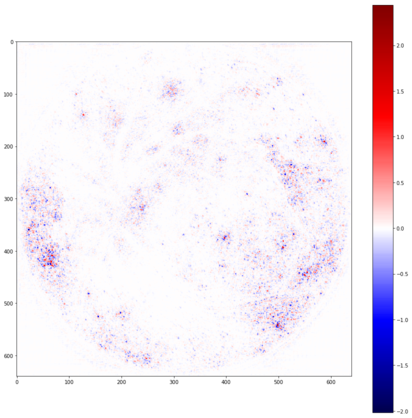}
		\caption{Total score for $C_3$}
		\label{fig:score_total_c3}
	\end{subfigure}
	\begin{subfigure}[b]{0.43\textwidth}
		\includegraphics[width=\textwidth]{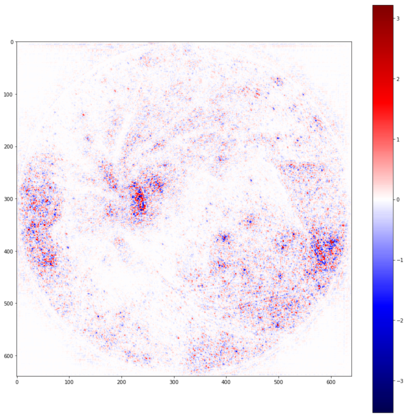}
		\caption{Total score for $C_4$}
		\label{fig:score_total_c4}
	\end{subfigure}~
	\begin{subfigure}[b]{0.43\textwidth}
		\includegraphics[width=\textwidth]{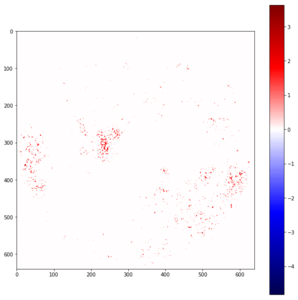}
		\caption{Total score for $C_4$ ($ S^{(in)} > 2\sigma$ )}
		\label{fig:score_total_c4_2std}
	\end{subfigure}
	\caption{Total score input-space distribution for classes 0, 1, 2, 3, 4 for test sample}
	\label{fig:test1_score_total}
\end{figure}

Score inputs can be combined with constant scores to define a unique input score map (see fig. \ref{fig:test1_score_total}). The sum of these scores is equal to the last layer inference score and determines the relative importance of every pixel in the final decision. A density plot and a standard deviation can also be calculated. In order to determine the importance of pixels is possible to restrict the visualization to positive scores or also be even more restrictive and visualize only pixels with a score greater than a predefined threshold, for example $n \sigma$ (see fig. \ref{fig:score_total_c4_2std}). These score maps are useful for building explanations, for detecting the cause of non-expected classifications, for example pixels with excessive importance in the final decision, conclusions based only on partial or incorrect information, etc.

\begin{figure}[h!]
	\centering
	\begin{subfigure}[b]{0.45\textwidth}
		\includegraphics[width=\textwidth]{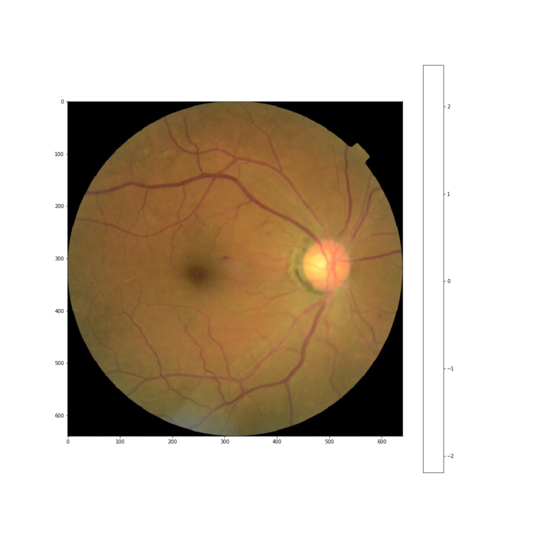}
		\caption{A $C_2$ sample}
	\end{subfigure}~
	\begin{subfigure}[b]{0.45\textwidth}		
		\includegraphics[width=\textwidth]{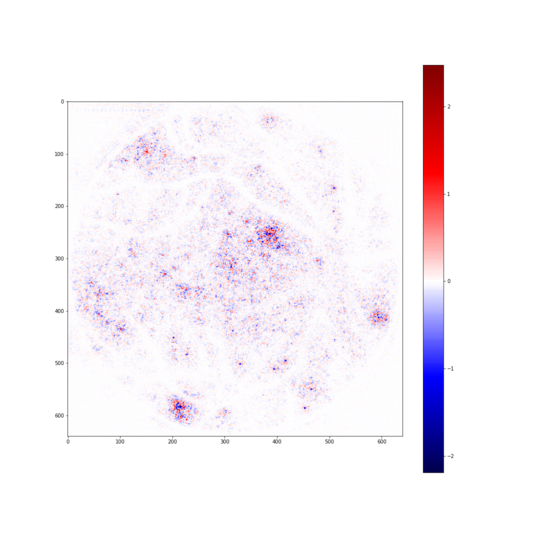}
		\caption{Score map generated for $C_2$ sample}
	\end{subfigure}
	\begin{subfigure}[b]{0.45\textwidth}
		\includegraphics[width=\textwidth]{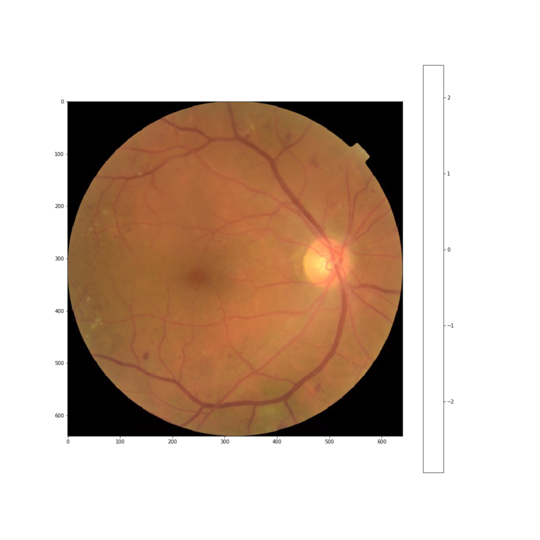}
		\caption{A $C_3$ sample}
	\end{subfigure}~
	\begin{subfigure}[b]{0.45\textwidth}
		\includegraphics[width=\textwidth]{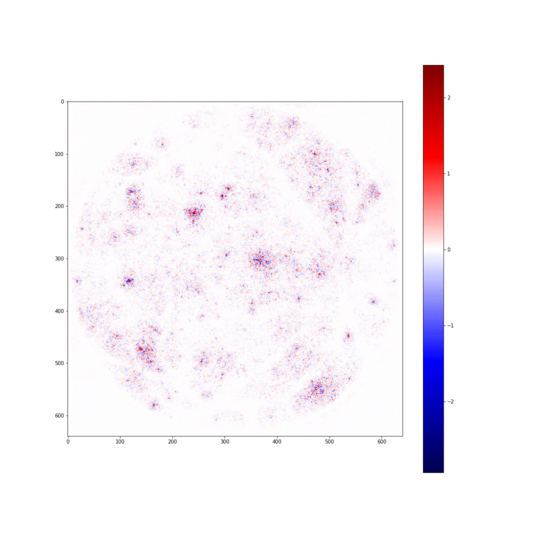}
		\caption{Score map generated for $C_3$ sample}
	\end{subfigure}
	\begin{subfigure}[b]{0.45\textwidth}
		\includegraphics[width=\textwidth]{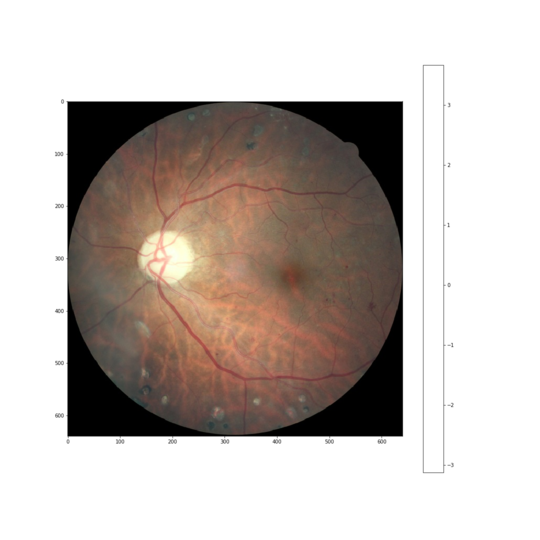}
		\caption{A $C_4$ sample}
	\end{subfigure}~
	\begin{subfigure}[b]{0.45\textwidth}
		\includegraphics[width=\textwidth]{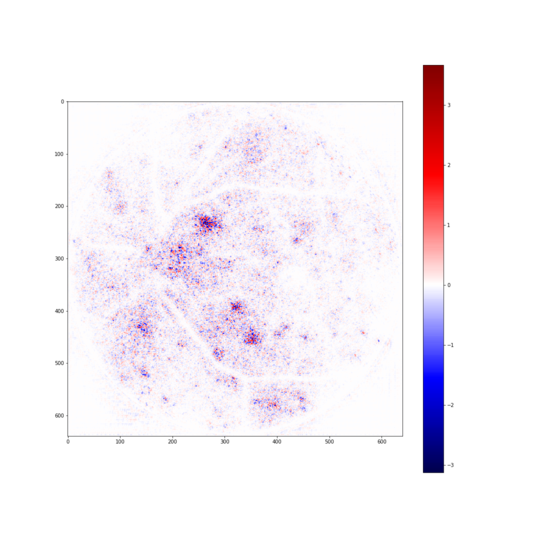}
		\caption{Score map generated for $C_4$ sample}
	\end{subfigure}

	\caption{Total score maps generated for different input images for the predicted class (class of maximum score)}
	\label{fig:score_samples}
\end{figure}

Fig. \ref{fig:score_samples} shows three different score maps generated for images belonging to diverse classes. For an appropriate analysis the score maps of every class should be considered and different threshold maps should be analyzed. In this figure due to space limitations only the predicted class map is shown. In future publications we will study the best method to extract conclusions of the generated maps.

\section{Conclusions}\label{sec:conclusions}

In this paper we presented a new model for the explanation of deep learning classification models based on the distribution of the last layer scores between the input pixels of the image. We presented a general theoretical derivation of the score calculation for the more typical deep learning building blocks to make possible the generation of score propagation networks for any type of applications based on deep learning models.  Additionally, we applied the model to design \emph{a human expert performance level DR interpretable classifier}. A model able to classify retine images into the five standardized levels of disease severity and able also to report, for every class, score importance pixel maps, providing the human expert the possibility of inference and interpretation. The score generation is done using a modified version of the pixel-wise relevance propagation algorithm, with the key difference of back-propagating only the part of the score that depends on the inputs and leaving the constant part as a contribution to the score of the considered layer. In this way, we are able to generate scores in a unique and exact way. Additionally, we developed a technique, consisting on applying a 2d-gaussian prior over the RFs, for mapping the constant hidden-space scores to the input, for generating a unique score map representative of the class, making possible to distribute the 100\% score class information of the last layer. 

\section*{Acknowledgements}
This work is supported by the URV grants 2014PFR-URV-B2-60 and 2015PFR-URV-B2-60, as well as, for the Spanish research projects PI15/01150 and PI12/01535 (Instituto de Salud Carlos III). The authors would like to thank to the Kaggle and EyePACS for providing the data used in this paper.

\section*{References}



\end{document}